\documentclass[a4paper]{article}
\usepackage{INTERSPEECH2014,amssymb,amsmath,epsfig}
\usepackage{algpseudocode}
\usepackage[toc,page]{appendix}
\usepackage{algorithm}
\ninept
\def\reg{{\rm\ooalign{\hfil
     \raise.07ex\hbox{\scriptsize R}\hfil\crcr\mathhexbox20D}}}

\title{Detection of texts in natural images}

\makeatletter
\def\name#1{\gdef\@name{#1\\}}
\makeatother \name{{\em Gowtham Rangarajan Raman\thanks{Work advised by Mr ES Gopi (Assistant Professor), NIT Trichy.}$^1$}}

\address{$^1$Department of Electronics and Communication Engineering, National Institute of Technology, Trichy, India \\
 {\small \tt 108108029@nitt. edu, gowthamrangarajan@gmail. com}}

\begin{document}
\maketitle
\begin{abstract}
A framework that makes use of Connected components and supervised Support machine to recognise texts is proposed. The image is preprocessed and 
and edge graph is calculated using a probabilistic framework to compensate for photometric noise. Connected components over the resultant image
is calculated ,which is bounded and then pruned using geometric constraints. Finally a Gabor Feature based SVM is used to classify the presence of text
in the candidates. The proposed method was tested with ICDAR 10 dataset and few other images available on the internet. It resulted in a recall and precision
metric of 0.72 and 0.88 comfortably better than the benchmark Eiphstein's algorithm. The proposed method recorded a 0.70 and 0.74 in natural images
which is significantly better than current methods on natural images. The proposed method also scales almost linearly for high resolution ,cluttered images.

\end{abstract}

\section{Introduction}

This report is divided into 3 sections ;Section 1  introduces to the problem and argues the need for a new framework. It concludes with presenting the proposed framework.\\
Section 2 describes the proposed framework ,It is further subdivided into subsections describing the modules the frame work uses. Each subsection argues the purposes of the module in the framework, the algorithm used and concludes with  a pragraph stating the exact values  of the parameters used in the implementation accompanying this report.\\
Section 3 discusses the Results and further improvements in the framework.\\
Readers who are interested in the a quick overview of the implemented method can read the last paragraphs of the corresponding sections and subsections. \\

\section{Introduction}

An essential prerequisite for text-based image search is the robust location of text within them. This is a challenging task due to the wide variety of variations in text appearances like font and style, geometric and photometric distortions, partial occlusions, and different lighting conditions. Text detection has been considered in many recent studies and numerous methods are reported in the literature \cite{epshtein2010detecting} \cite{sumathi2012survey}. These techniques can be classified broadly into two categories: texture-based and connected component (CC)-based.\\

Texture-based approaches view text as a special texture that is distinguishable from the background. Typically, features are extracted over a certain region and a classifier (trained using machine learning techniques or by heuristics) is employed to identify the existence of text. In [11], Zhong et al. assume that text has certain horizontal and vertical frequencies and extract features to perform text detection in the discrete cosine transform domain. Ye et al. collect features from wavelet coefficients and classify text lines using SVM \cite {Ye2005textdetection}. Chen et al. feed a set of weak classifiers to the Adaboost algorithm to train a strong text classifier \cite {chen2004detecting} \cite{chen2005time}.\\

Methods using machine learning are computationally intensive since they compute features for all sections of the image at different resolutions using a moving window of varying size. However the maturity of learning algorithms further the chances of figuring out the best representation of an image which could be used for text detection problems. For example, Convlutional Neural Network has been trained recently by reCaptcha Team at google \cite {goodfellow2013multi} to classify regularized texts in doors with 99+ \%/  accuracy and the computational power required to achieve the result renders any ordinary computer to shame. \\

The CC-based approach extracts regions from the image and uses geometric constraints to rule out non-text candidates. Adaptive binarization are usually used to find CCs. Text lines are then formed by linking the CCs based on geometric properties. Recently, Epshtein et al. \cite {epshtein2010detecting} proposed using the CCs in a stroke width transformed image, which is generated by shooting rays from edge pixels along the gradient direction. Shivakumara et al. extract CCs by performing K-means clustering in the Fourier-Laplacian domain, and eliminate false positives by using text straightness and edge density \cite{shivakumara2011laplacian}. \\

CC based methods are computationally inexpensive since they reject most of the candidates before scrutinizing for text information. Hard pruning necessitates  the graph constructed,on which Connected components are formed, [edge graph] to be robust towards lighting changes , clutter etc., These methods completely discard texture information pertaining to text making them inefficient in cluttered backgrounds. Texture 'similarity' within a  group of text is also discarded in these methods , which will make a huge difference in poor resolution pictures.\\

Modelling texts against non-texts are problems of high dimensions due to the sheer number of poses , fonts , textures and other possibilities of texts occurring in natural images, tackling them using low-dimensional models may not improve our chances of success. Hence , we would be treating the problem in high dimensions leveraging machine learning  techniques to solve the problem.\\
Supervised Machine learning classifications  are computationally intensive. The complexity of the method is predominantly not due to its complexity involved in transforming and classifying the region of interest but in the number of candidates on which the classification rule is run. it is general procedure to generate candidates by sliding a window of fixed dimensions over the entire image and on its scaled form. This procedure generates huge number of candidates exponentially increasing the complexity of the algorithm. Instead of considering possibility of texts in all possible locations in an image ,one could apply CC based method to narrow down the number of possible candidates to be considered and then apply machine learning methods on these regions to classify. This reduces the complexity by leaps and bounds. This route is taken in this report to solve the problem.\\

\section{Framework}
In simple terms the problem could be defined as following.
Given an image , isolate text components in separate bounding box as highlighted in yellow below \\
\begin{figure}[htb]
\begin{minipage}[b]{1.00\linewidth}
  \centering
  \centerline{\includegraphics[width=9.4cm]{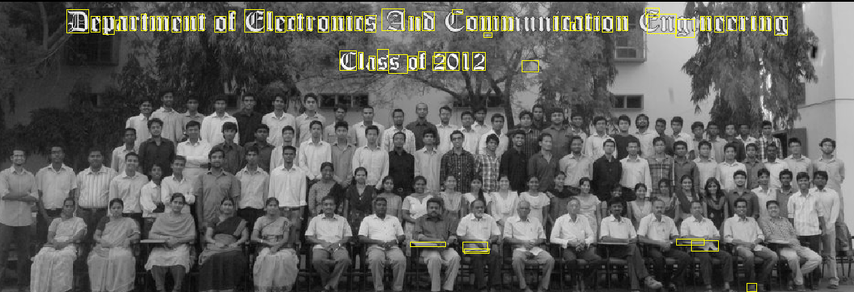}}
 \vspace{-0.4cm}
  %\centerline{Possible textual content in this synthesised image is bounded by yellow boxes by the proposed algorithm}\medskip
\caption{Possible textual content in this image is bounded by yellow boxes by the proposed algorithm.}
\label{fig:fig1}
\end{minipage}
\end{figure} 
The image is initially pre processed to remove distortion due to lighting, increase sharpness through histogram equalization methods and sharpening filters. An edge graph of the resulting image is constructed based on Marimonts probabilistic framework \cite {marimont1998probabilistic} which computes the probability of the presence of an edge at a certain pixel  modelling noise due to photometric distortion. This edge map is segmented into distinct regions based on its nearest neighbour using breadth first search. The independent regions,which are possible text candidates, are bounded using rectangular boxes. These candidates are then discarded based on geometric constraints specific to english texts. Few candidates are  also discarded based on extent of overlap. Finally, Gabor features \cite {zhang2002robust} of  remaining candidates are calculated and passed through an SVM classifier to assert the presence of text. A concise flow of the algorithm explained is presented in the form of blocks of pseudo code and the details of which are explained in the succeeding pages. The framework is explained in detail in the succeeding paragraphs.

%%%%%%%%%%%%%%%%%%%%%%%%%%%%%%%
\vspace{-2mm}
\begin{algorithm}[H]
  \caption{Overview of the process followed for detecting texts}
  \begin{algorithmic}
    \State  Pre-process the Image \textbf{I} through \textbf{S},Sharpening filter and  \textbf{H}, Histogram equalization filter to obtain \textbf{O}
	\State  Generate Edge map \textbf{E} using Marimont's framework on \textbf{O} 
	\State  Segment \textbf{E} using Connected component to $\textbf{P}_i$	
	\hspace{10mm}
	
    \For  {$i=1$ to $ELEMENTS$} 
	\State  Calculate Gabor feature $\textbf{G}_i$ on resized Image $\textbf{P}_i$	
	\State  Classify $\textbf{G}_i$  using trained SVM 
	\EndFor 
	
  \end{algorithmic}
\end{algorithm}
\vspace{-2mm}
%%%%%%%%%%%%%%%%%%%%%%%%%%%%%%%

\subsection{EdgeMap}
\vspace{-2mm}
The input image is preprocessed to increase sharpness using Matlab’s built in Imsharpen method which uses a Laplacian Kernel. Lighting distortion  is tackled by using local histogram equalization methods. Contrast Local Adaptive Histogram Equalization method which is present in Matlab \cite{zuiderveld1994contrast} was used to offset poor lighting and contrast. The preprocessed image is used to create an edgemap.

The edge map forms the crux of detecting textual candidates. Hence a lot of importance is placed to ensure that the graph generated captures the details and remains robust during poor lighting. A canonical filter like sobels or canny would  undermine the potential of the proposed algorithm to a large extent in natural scenes. Nevertheless, they work just fine in digitally born images where clutter is the main problem and not poor lighting. To tackle the uncertainty of lighting in natural scenes we use  the statistical framework proposed by Marimont and Rubner \cite {marimont1998probabilistic} and generate the edge map instead of a spontaneous second derivative canny edge. \\

The edge map is calculated from two probability maps , a zero crossing map  and a confidence map. The presence of an edge at a pixel can be either due to noise in the image or the presence of edge at the position. The statistical approach works significantly well in capturing minute edge details in natural images. The method is chosen for its robustness and its successful application in many low-level image processing algorithms \cite {rosenholtz2000significantly} \cite {neoh2004adaptive} \cite {i2005sparse}.\\

The edge map algorithm models the input signal S(x,y) as the sum of true signal R(x,y) and additive gaussian noise  $\\etta(x,y)$. The zero crossing edge map  is generated by applying the directional Lindberg operator on the raw image which is a modified bessel function. The second and third derivative of the image calculated using the Lindberg operator \cite {lindeberg1993scale} is used reject phantom edges. The probability of a zero crossing is written as an integral sum as in equation,where the conditional probability of the presence of edge given directional derivative is obtained using the Lindberg operator. For our purposes we replace the integration with summation over fixed intervals. The conditional probability is taken to be a normal distribution with unit variance around the gradient of highest descent. \\

\begin{align}
\textbf{p}(edge|\theta,r) = G(\frac{\sigma_{3}}{\sigma_{2}} f(\delta x)) - G(\frac{-\sigma_{3}}{\sigma_{2}} f(\delta x))\\
\textbf{p}(edge|\theta,r) = G(\frac{\sigma_{3}}{\sigma_{2}} f(\delta x)) - G(\frac{-\sigma_{3}}{\sigma_{2}} f(\delta x))\\
\textbf{G}(t) = L(\frac{a-bt}{1-t^{2}}-b,\frac{t}{1-t^{2}}) + L(\frac{-a+bt}{1-t^{2}}+b,\frac{t}{1-t^{2}})
\end{align}

The confidence map that would be generated measures the odds that the edge detected is due to true signal’s zero crossing and not due to the noise. The fundamental notion of estimating the confidence of a signal with additive gaussian noise as discussed by Murihead \cite {muirhead2009aspects} has been borrowed.  A confidence Kernel k(n) is constructed around each pixel, which is then used as parameters for a $\chi ^{2}$ distribution. The confidence of the presence of an edge is measured as the top 2% of the $ \chi ^{2}$ distribution with the parameters obtained. Refer \cite {marimont1998probabilistic} for more.\\

The implementation of the algorithm here assumes the variance of the sensor noise as 1. It uses a 3X3 Lindberg operator varying its direction in steps of 30 degrees. The presence of an edge at a pixel is calculated as a weighted sum of confidence and the edge density along  8 adjacent neighbours. If their mean exceeds 0.5 then an edge is assumed to be present.\\
To summarize, the edgemap generation algorithm involves calculating kernel matrix and calculating the top 2 \% of the $ \chi^{2}$  distribution for generating the confidence map ;  calculating the directional lindberg derivative over 12 directions per pixel to compute presence of zerocrossing;.Though complexity is asymptotically linear , it may behave quadratically for images of smaller size. The edgemap this generated are used to extract regions of interest.\\
\subsection{Prunning}
The edgemap representation of the image is segmented into connected components. Specifically ,a Breadth first search to analyze 4 connected neighbors ,in the edge map, is performed dividing the edge map into a labeled graph. Then,for each disconnected graph center , maximum height and width that encompasses its component are calculated and the distinct connected region is bounded by a rectangular box. Candidates [bounding-box] having Euler number < 2 , and those having an area of <2% of image area are removed immediately. The Candidates which overlaps more 80% with others are removed such that the area covered by all the remaining candidates put together is maximum. A dynamic programming solution is used to tackle this problem. Clearly the preprocessing techniques based on the number of candidates from edgemap has linear complexity and is much faster than other windowing based techniques performed on the entire image. Gabor features are extracted from the remaining candidates.\\

\subsubsection{Feature Extraction }
Gabor Wavelet is a sinusoidal plane wave with particular frequency and orientation modulated by a Gaussian envelope. It is a filter of the following equation

\begin{align}
	h(x,y) = e^{(\frac{x^{2}}{\sigma_{x}^{2}}+\frac{y^{2}}{\sigma_{y}^{2}})}e^{-j2\pi(ux+vy)}
\end{align}

Gabor Filters, which operate directly on gray-level images,have several advantages \cite {Ye2005textdetection}. First, Gabor features have been used for capturing local information in both spatial and frequency domains from images, as opposed to other global techniques such as Fourier Transforms. The Filtering output is robust to various noises since G. However the orientation specificity of Gabor wavelets force us to use filters in different directions adding to the complexity. The major motivator comes for success in using gabor filters for recognizing  texts \cite {jain1992text} \cite {zhang2002robust}.\\

In the subsequent filtering and isolation of text candidates using SVM learning method , we extract the gabor filter features for each bounding box. Features are obtained by convoluting a 3*3 gabor matrix  over the candidate region. The resultant region is resampled to 20*20 block. Four Gabor filters along different orientations, by variying $\sigma_{x}$, $\sigma_{y}$ equally,  are applied to the input image leading to 4, 20*20 features per candidate. These were reduced to fewer dimensions along Principal components. PCA for  these features were calculated and the top 20 (\~30\%) were retained . It was then classified using trained SVM classifier written. Python was used for Spectral decomposition \cite {scikit-learn}.\\

\subsubsection{Training}
Support Vector Machine (SVM) learner/classifier is used in the problem. The idea behind Binary SVM is to dichotomize the given dataset using a line or find a line that shatters the dataset as maximally as possible. Clearly the idea assumes linear separability of the dataset ,which might not be the case with most data, certainly not with text and non-text data. A kernel trick ,using a non-linear error measure,makes it possible to use SVM for non-linear data too. The theory behind SVMs are not discussed for the sake of brevity. Extensive discussions on SVM can be found in \cite {marimont1998probabilistic}.\\

The SVM with gabor features as input was trained on a database consisting of 600 samples labeled as text obtained from ICDAR datasets and 200 other manually labelled natural texts using the sci-kit learn package \cite {wolf2006object}. Non-text labelled data was also collected from the same sample. Radial basis function kernel to train the SVM where the normalised kernel bandwidth \textbf{s} was determined through cross-validation among \textbf{s} with a granularity of 0.1 from -1 to 1. The python machine learning library scikit  learn was used to implement the algorithm \cite {scikit-learn}. \\

\section{Results and Discussion}

\subsection{Complexity}
The complexity of the framework is linear. The code was written for readability and re usability ,hence critical parts were not written in languages like C/C++  (that generates fast assembly code) compromising the efficiency of implementation. The approximate time taken to solve the problem against the no of pixels used are provided below for objective comparison. The algorithm ran in a 32 bit Matlab 2011R b (64 bit Windows OS) using Core i5 4GB ram 2.4Ghz  Dell Inspiron Laptop. The SVM classifier ran on python’s scikit learn library. The SVM classifier training took 50 minutes using the same laptop
\begin{figure}[htb]
\begin{minipage}[b]{1.00\linewidth}
  \centering
  \centerline{\includegraphics[width=9.4cm]{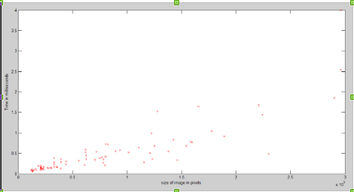}}
 \vspace{-0.4cm}
  \caption{Time taken in millisec vs Number of Pixels. The graph indicates asymptotic linearity in complexity for most images.}
\label{fig:fig2}
\end{minipage}
\end{figure} 

\subsection{Robustness}
The goal of a rectangle based object detection evaluation scheme is to match a list  of ground truth object rectangles $\textbf{G}_i$  and a list D of detected object rectangles $\textbf{D}_i$ and to measure the quality of the match between the two lists. The quality measure should penalize information loss, which occurs if objects or parts of objects have not been detected, and it should penalize information clutter, i. e. false alarms or detections which are larger than necessary. Precision and recall metrics are used to evaluate measure the matching. Intuitively , recall is the ratio between regions of ground truth that has been correctly detected and the ground truth. Precision is the amount of false positive area being detected by the algorithm. These metrics are mathematically defined for matches between one ground truth and one detected region as shown in the equation.
\begin{align}
 R_{ar} = \frac{Area(G_{i}\bigcap D_{i})}{Area(D_{i})}\\
 P_{ar} = \frac{Area(G_{i}\bigcap D_{i})}{Area(G_{i})}
 \end{align}
This definition is extended by Wolf et., al \cite {wolf2006object} for matching multiple ground truth with multiple detected regions as shown below. This handles the case of a  subset of detected regions cover  single ground truth much better than the earlier definition but not as completely as desired.
\begin{align}
 BestMatch_{D} = max(\frac{Area(G_{i} \bigcap D_{i})}{Area(G_{i} + Area(D_{i})}) , i \in 1,2,...|D|\\
 BestMatch_{G} = max(\frac{Area(G_{i} \bigcap D_{i})}{Area(G_{i} + Area(D_{i})}) , i \in 1,2,...|G|\\
 P_{ICDAR}(G,D) = \frac{\sum_{i=1}^{|G|}BestMatch_{D}(G_{i})}{|D|}\\
 R_{ICDAR}(G,D) = \frac{\sum_{i=1}^{|D|}BestMatch_{G}(D_{i})}{|G|}
 \end{align}

The robustness results are defined based on metrics defined. Precision measures ability the minimality of false positives. Recall measures the minimality of true negatives. The overall metric favours lower Recall rates  non linearly over high precision. The algorithm as expected was able to detect all text regions , However it also misclassified  non-text regions more often than expected.

\begin{figure}[htb]
\begin{minipage}[b]{1.00\linewidth}
  \centering
  \centerline{\includegraphics[width=8.0cm]{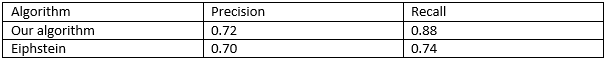}}
 \vspace{0.4cm}
 
  \caption{Comparison Table }
\label{fig:fig3}
\end{minipage}
\end{figure}

Eiphstein's algorithm and ours were compared using the validation test. The algorithm was tested in IDCAR database and encouraging results were obtained. It was also tested on natural images and the results were encouraging too. The exact details on IDCAR dataset are provided below

\begin{figure}[htb]
\begin{minipage}[b]{1.00\linewidth}
  \centering
  \centerline{\includegraphics[width=9.4cm]{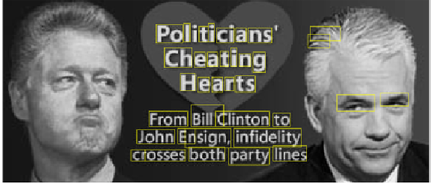}}
 \vspace{0.4cm}
 \centerline{\includegraphics[width=9.4cm]{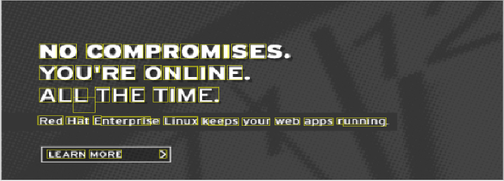}}
 \vspace{0.4cm}
  \caption{Results observed in natural and digital images}
\label{fig:fig3}
\end{minipage}
\end{figure} 

\subsection{Observation and Improvements}
There were few observations about the algorithm which can be considered to improve the quality of the output generated. Labelled Negative samples were not sufficiently ‘confusing’ for the SVM This is clearly discernible since texts were mostly confused with dense dark strokes -twigs and eyes share. It was also found that careful coalescing of region of interests which are ‘close-enough’ would significantly improve the performance of the system , since Isolated texts are hard to distinguish than word strings. For eg., ‘I’,independently can be easily confused with other dark strokes ,on the contrary a string of letters are clearly distinguishable. A  Spanning Tree based merging followed by pruning could be a possible way to handle them. It was implemented as an extension.\\

The strongest motivation behind choosing SVM over other learning methods is because of its unmatched results in Image processing and its scope for incremental learning. Support Vector Machines could be modified to suit reinforced learning structure which would lead to a more robust learning framework evolves over time \cite {laskov2006incremental}.

%%%%%%%%%%%%%%%%%%%%%%%%%%%%
%\begin{appendix}
%{
%\section{ Appendix 1} \label{App:AppendixA}
%\begin{align}
%\textbf{K}_k&=\text{argmin }\mathbb{E}\big [(\textbf{s}_k-\hat{\textbf{s}}_k)^T(\textbf{s}_k-\hat{\textbf{s}}_k)\big ]\nonumber \\
%&=\mathbb{E}\Big [\big (\textbf{s}_k-\{\textbf{s}_k^-+\textbf{K}_k(\textbf{x}_k-\textbf{H}\textbf{s}_k^-)\}\big )^T\nonumber\\
%&\hspace{10mm}\big (\textbf{s}_k-\{\textbf{s}_k^-+\textbf{K}_k(\textbf{x}_k-\textbf{H}\textbf{s}_k^-)\}\big )\Big ]\nonumber\\
%&=\mathbb{E} \Big [ \big ( (\textbf{s}_k-\textbf{s}_k^-)-\textbf{K}_k\textbf{H}(\textbf{s}_k-\textbf{s}_k^-)\big )^T\nonumber\\ &\hspace{10mm}\big %((\textbf{s}_k-\textbf{s}_k^-)-\textbf{K}_k\textbf{H}(\textbf{s}_k-\textbf{s}_k^-)\big )]\nonumber\\
%&=\mathbb{E} \Big [ \big ( (\textbf{s}_k-\textbf{s}_k^-)-\textbf{K}_k\textbf{H}(\textbf{s}_k-\textbf{s}_k^-)\big )^T\nonumber\\ &\hspace{10mm}\big %((\textbf{s}_k-\textbf{s}_k^-)-\textbf{K}_k\textbf{H}(\textbf{s}_k-\textbf{s}_k^-)\big )]\nonumber\\
%\intertext{Differentiating  and equating to zero gives}
%&\mathbb{E}\Big [ %(\textbf{s}_k-\textbf{s}_k^-)%(\textbf{s}_k-\textbf{s}_k^-)^T\textbf{H}^T+\nonumber\\&\hspace{20mm}\textbf{K}_k\textbf{H}%(\textbf{s}_k-\textbf{s}_k^-)%(\textbf{s}_k-\textbf{s}_k^-)^T\textbf{H}^T\Big]=0\nonumber\\
%\textbf{K}_{k}&=\textbf{P}_{k}^{-}\textbf{H}^{T}(\textbf{H}\textbf{P}_{k}^{-}\textbf{H}^{T})^{-1}\nonumber
%\end{align}
%}
%\end{appendix}

\eightpt
\bibliographystyle{IEEEtran}
%\begin{thebibliography}{10}
%\bibliographystyle{ieeetr}
\bibliography{ref}
%\end{thebibliography}
\end{document}